\documentclass[lettersize,journal]{IEEEtran}
\usepackage{amsmath,amsfonts}
\usepackage{algorithmic}
\usepackage{algorithm}
\usepackage{array}
\usepackage[caption=false,font=normalsize,labelfont=sf,textfont=sf]{subfig}
\usepackage{textcomp}
\usepackage{stfloats}
\usepackage{url}
\usepackage{verbatim}
\usepackage{graphicx}
\usepackage{cite}
\begin{document}

\title{Path Planning based on 2D Object Bounding-box}

\author{Yanliang Huang, Liguo Zhou, Chang Liu, Alois Knoll,~\IEEEmembership{Fellow,~IEEE}
\thanks{All authors are with the Chair of Robotics, Artificial Intelligence and Real-time Systems, Technical University of Munich, Parkring 13, 85748 Garching, Germany {\tt\small liguo.zhou@tum.de knoll@in.tum.de}}
}



\maketitle

\begin{abstract}
The implementation of Autonomous Driving (AD) technologies within urban environments presents significant challenges. These challenges necessitate the development of advanced perception systems and motion planning algorithms capable of managing situations of considerable complexity. Although the end-to-end AD method utilizing LiDAR sensors has achieved significant success in this scenario, we argue that its drawbacks may hinder its practical application. Instead, we propose the vision-centric AD as a promising alternative offering a streamlined model without compromising performance. In this study, we present a path planning method that utilizes 2D bounding boxes of objects, developed through imitation learning in urban driving scenarios. This is achieved by integrating high-definition (HD) map data with images captured by surrounding cameras. Subsequent perception tasks involve bounding-box detection and tracking, while the planning phase employs both local embeddings via Graph Neural Network (GNN) and global embeddings via Transformer for temporal-spatial feature aggregation, ultimately producing optimal path planning information. We evaluated our model on the nuPlan planning task and observed that it performs competitively in comparison to existing vision-centric methods.
\end{abstract}


\section{Introduction}
\IEEEPARstart{R}{ecent} advancements in deep learning have significantly transformed the field of AD \cite{zhang_attention-based_2023, hu_planning-oriented_2023, hu_st-p3_2022, vitelli_safetynet_2021}. The traditional pipeline of AD systems is structured around three core stages: perception, prediction, and planning \cite{teng_motion_2023}. These stages have traditionally been constructed through manual engineering approaches \cite{gong_real-time_2023}, which often face challenges in terms of scalability and addressing complex, less frequent scenarios (long-tail problems). The perception and prediction stages of AD have seen considerable improvements with the advent of machine learning. However, the incorporation of machine learning into the planning stage presents a complex challenge, primarily due to the intricate nature of real-world decision-making processes. This paper seeks to address these challenges by exploring the application of deep learning in a comprehensive manner across the entire spectrum of AD, advocating for an end-to-end learning approach.

The perception systems in AD vehicles predominantly diverge into two categories: LiDAR-based and camera-based. These methods are fundamentally reliant on the detection and recognition of objects, utilizing 3D point clouds and 2D images respectively. Despite extensive discussions and comparisons between these two approaches, a definitive resolution has not yet been achieved. LiDAR sensors, while effective in certain aspects, display susceptibility to adverse weather conditions \cite{hasirlioglu_test_2016} and involve intricate moving components. Furthermore, they pose potential risks to the CMOS and CCD sensors used in electronic cameras and the environment, due to laser-induced damage \cite{schwarz_laser-induced_2017}. These factors make LiDAR less suitable for large-scale deployment. On the other hand, vision-centric systems, which exclusively depend on sets of high-resolution cameras for environmental perception, are gaining traction as not only cost-effective in capturing rich visual information but also more aligned with sustainable development goals. This shift towards vision-centric solutions is increasingly seen as a potential future direction for the AD sector. In light of these considerations, our research focuses on leveraging camera systems to develop an AD method that takes images and videos as input.

\begin{figure}[t]
\centering
\includegraphics[width=0.5\textwidth]{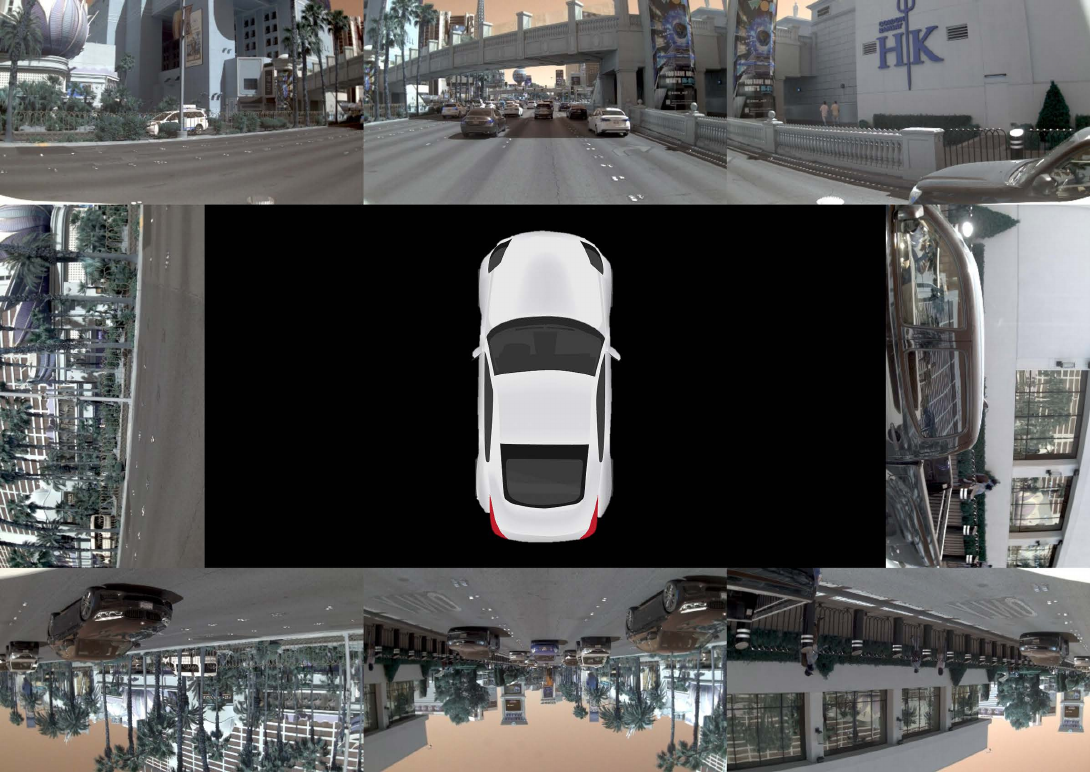}
\caption{An example showing a vehicle equipped with multiple cameras capturing the surrounding environment.}
\label{fig_5}
\end{figure}

The YOLO (You Only Look Once) \cite{redmon_you_2016} algorithm, renowned for its lightweight and efficient design, has demonstrated significant success in object detection and tracking \cite{liu_yolo-bev_2023}. Its ability to swiftly identify and track objects using car cameras positions YOLO as an effective real-time perception method. In this study, we utilize a pre-trained YOLO model as the perception backbone, generating bounding-box detection and object-tracking information from multi-camera data in various driving scenarios. However, camera sensors are prone to environmental influences and partial perception issues, presenting challenges in autonomous driving. To address these, we integrate HD maps of city areas and data from various on-board sensors to obtain a more comprehensive dynamic understanding of the driving environment.

In the planning phase, our approach incorporates a dual embedding attention-based GNN algorithm inspired by \cite{gao_vectornet_2020, scheel_urban_2021}. This method locally and globally aggregates driver information to predict the future pathways of the vehicle, effectively learning the mapping from perception outputs, HD maps, and sensor data to the vehicle's final actions.

Our model, structured around these deep learning-based stages, provides a simple, lightweight, and rapid solution for real-time decision-making in complex urban environments. Leveraging large-scale datasets, our system is capable of mimicking the human-driver decision-making process in intricate situations by generating future waypoints, which can then be further translated into control signals for the vehicle. We believe our approach not only complements existing methods but also opens avenues for future research in this domain, particularly in the development of Level 5 automated systems. To sum up, key contributions of this work include:

\begin{itemize}
    \item[$\bullet$] Proposing a lightweight model to address real-time urban driving challenges.
    \item[$\bullet$] Modeling the interplay between vision-based sensing, HD maps, and dynamic data for contextual understanding.
    \item[$\bullet$] Implementing local and global embedding to predict driving actions effectively.
    \item[$\bullet$] Demonstrating rapid response and superior performance of our model in the nuPlan challenge, setting a benchmark for future developments in AD.
\end{itemize}

\section{Related Works}
This section provides an overview of various methodologies employed in AD decision-making, evaluating their respective outcomes and limitations, with a specific focus on the techniques in the perception and planning stages.

\subsection{Pipeline Planning Methods}
In contemporary AD development, pipeline planning algorithms are prevalent, primarily due to their interpretability and straightforward structure. These methods typically encompass distinct modules such as perception, localization, planning, and control, each constructed using conventional techniques. Notably, the Kalman Filter is often employed for sensor fusion, and the A* search algorithm \cite{ziegler_navigating_2008}, is utilized for path planning. However, these methods predominantly rely on hand-crafted feature engineering and manually defined functions, which present significant challenges when processing large-scale data. This limitation becomes particularly evident in urban driving scenarios, where achieving safe and efficient driving control is crucial. The traditional approach often struggles with the complexity and unpredictability inherent in these environments. Additionally, these methods are susceptible to accumulative errors and may exhibit deficiencies in task coordination. This susceptibility is mainly due to the segmented nature of the pipeline approach, where errors in one module can propagate and amplify through subsequent stages.

In our research, we adopt the perception-planning framework as the foundational structure for our model. However, unlike traditional pipeline planning methods, we integrate machine learning-based techniques to enhance performance. This integration aims to address and effectively navigate long-tail problems, which are scenarios that are infrequent yet critical in the context of urban AD. By leveraging the capabilities of machine learning, our approach seeks to surmount the limitations of conventional pipeline planning methods, particularly in terms of adaptability and responsiveness to the dynamic and diverse challenges encountered in urban driving environments. This paradigm shift in methodology underscores the potential of machine learning in revolutionizing the field of AD, paving the way for more robust, adaptive, and safe autonomous vehicles.

\subsection{Learning-Based Method}
Addressing the drawbacks of modular methods, the trend in AD research has shifted towards directly mapping raw perception data to control commands, known as the learning-based or end-to-end method \cite{grigorescu_survey_2020}. This approach unifies all aspects—from perception and prediction to planning—into a cohesive whole, optimizing the entire driving task through a single gradient descent process in the network. Automated learning of feature representations from high-dimensional inputs (such as images and point clouds) allows the model to intuitively grasp driving principles, potentially surpassing human-defined methods in handling long-tail problems and achieving robustness.

However, the lack of intermediate, interpretable representations in learning-based methods poses challenges in explicating the rationale behind specific command executions in given scenarios. This limitation is particularly concerning regarding safety in dynamic urban driving environments.

Our proposed solution merges the principles of pipeline planning with learning-based approaches. We divide our model into two independent yet interconnected learning-based modules: perceiving surrounding objects and explicitly planning a series of safe waypoints. This approach enables our model to benefit from the high performance of learning-based methods while retaining a degree of interpretability intrinsic to module-based methodologies. Through this combination, we aim to strike a balance between performance enhancement and the need for clarity and safety in decision-making within AD systems.

\subsection{Object Detection Methods in Perception}
Perception, greatly advanced by modern deep learning methods, encompasses numerous techniques for image analysis. Object detection methods can be categorized into two types: two-stage and one-stage detectors. Two-stage detectors, such as the R-CNN family (from vanilla R-CNN \cite{girshick_rich_2014} to Mask R-CNN \cite{he_mask_2017}), demonstrate impressive performance in image detection and segmentation but suffer from high computational costs due to object proposal stages. In contrast, one-stage detectors like YOLO \cite{redmon_you_2016}, SSD \cite{liu_ssd_2016}, and RetinaNet \cite{lin_focal_2017}, forgo object proposals, offering faster operation with competitive accuracy. This speed is particularly advantageous in handling the crowded and dynamic conditions of real-time urban driving. YOLO, evolving rapidly over the years, provides unparalleled performance in terms of speed and accuracy. In our model, we employ YOLOv8 as the backbone of our perception network, capitalizing on its strengths to address the complexities of urban AD scenarios.

\subsection{Path Planning Methods}
In the sphere of machine learning-based path planning for AD, reinforcement learning \cite{levine_continuous_2012}, and imitation learning \cite{codevilla_end--end_2018} have emerged as the most influential and promising approaches, supported by a substantial body of pioneering research. In reinforcement learning, the learning agent (reinforcer) acquires appropriate actions for given states by engaging in a process of continuous exploration and optimization of its policy, facilitated through a deep learning network. This methodology inherently requires the use of a simulator and a well-defined, explicit reward signal to guide the learning process. However, a significant challenge with these approaches is their propensity to overfit to their training environments, leading to suboptimal performance in real-world, long-tail scenarios. To mitigate this, our approach employs a simulator constructed from a comprehensive collection of real-world driving logs \cite{jain_autonomy_2021}, thereby diversifying the training environment, and enhancing the model's ability to generalize to a wider range of real-world conditions.

Imitation learning represents a nuanced form of reinforcement learning, where the learning agent acquires knowledge from expert demonstrations, striving to replicate the behavior of a human driver. With a sufficient volume of training data, the learner is enabled to emulate human-like driving behavior and decision-making processes. Within imitation learning, the implementation of various neural network architectures, particularly GNN, has been noteworthy. A critical aspect of this approach is the effective assimilation of extensive and diverse data derived from urban driving scenarios. In this context, VectorNet \cite{gao_vectornet_2020}, is an efficient approach utilizing a hierarchical GNN. VectorNet initially leverages the spatial locality of individual road components, represented as vectors, and subsequently models the intricate interactions among these components. This method achieves exceptional performance while maintaining a streamlined network architecture, significantly reducing the number of parameters. This efficiency is particularly crucial for our objective of developing a real-time urban driving planning method. In our research, we have adapted and fine-tuned VectorNet to align with the specific requirements of our path planning model.

\begin{figure*}[!t]
\centering
\includegraphics[width=0.8\textwidth]{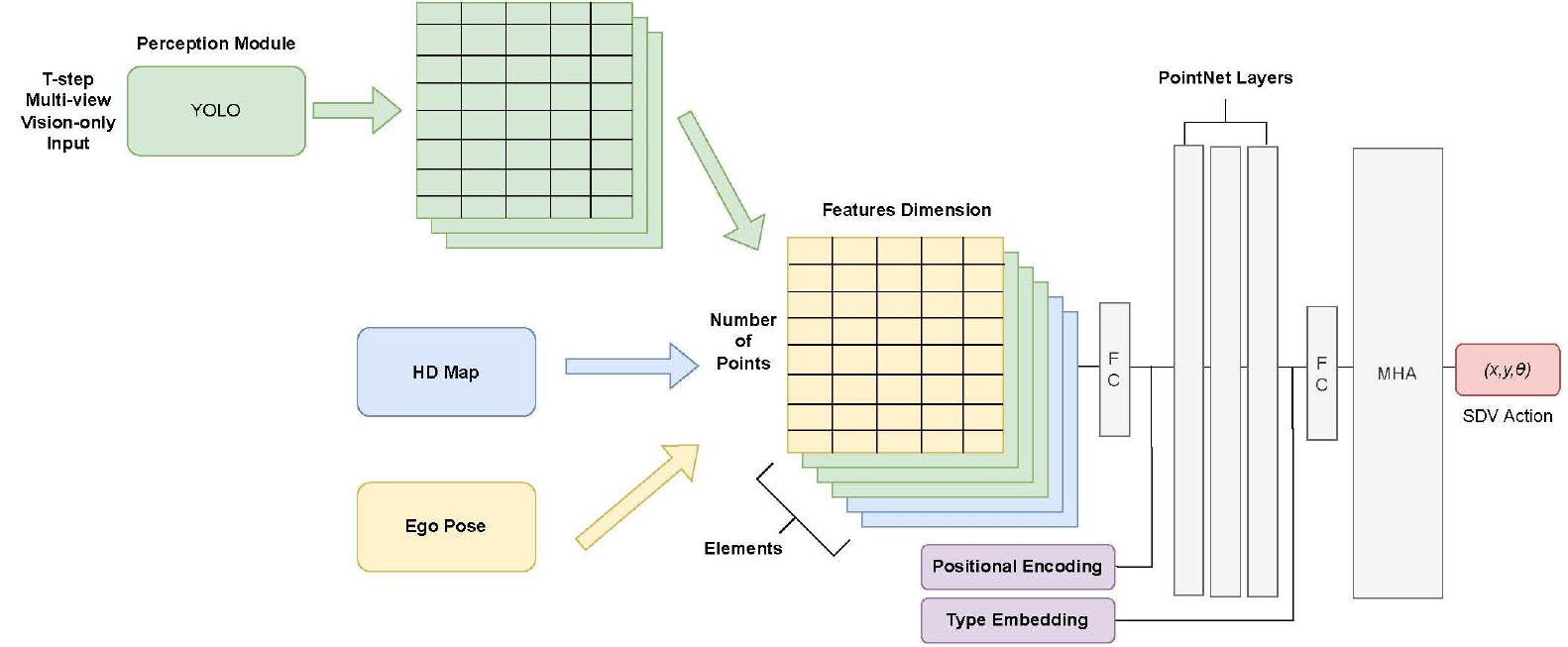}
\caption{General framework of the proposed network, For perception, the YOLO backbone ensures fast and reliable bounding-box detection and tracking. For planning, combined with HD Map and Ego Pose, the prior knowledge is fed into a dual-embedding GNN to generate the final trajectory.}
\label{fig_1}
\end{figure*}

\section{Methodology}

\begin{figure*}[t]
\centering
\includegraphics[width=0.8\textwidth]{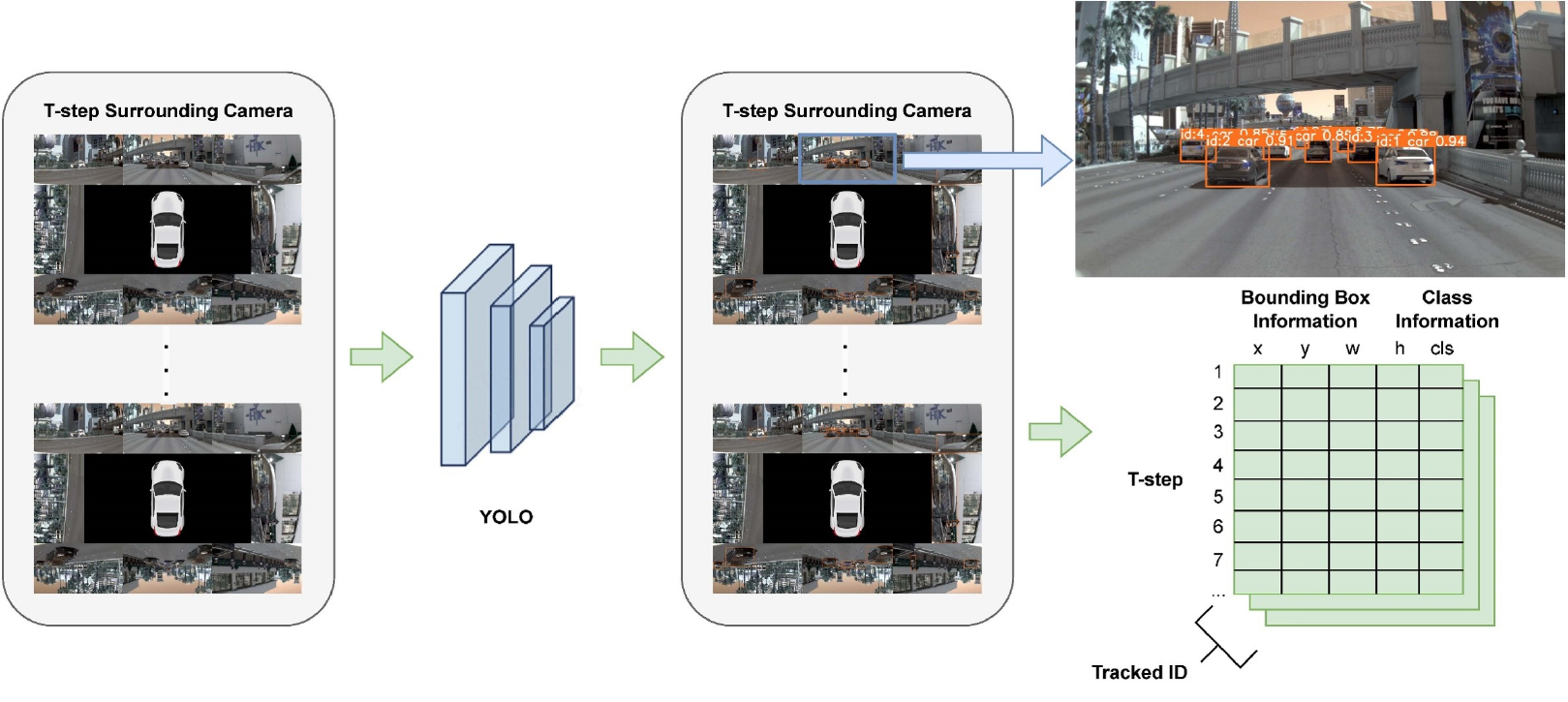}
\caption{
The image matrix is input into a YOLO backbone, from which bounding box coordinates, class identifications, and tracking information are generated. This process facilitates the accurate detection and classification of agents within the image, as well as their tracking over time.}
\label{fig_2}
\end{figure*}

Our system is designed to analyze video frames captured from multi-view on-board cameras, with the objective of predicting feasible maneuvers for autonomous driving. We treat it as a regression problem. The entire framework is depicted in Figure \ref{fig_2}. Initially, the YOLO algorithm processes the video frames to extract features from the surrounding environment and generates bounding boxes with corresponding tracking information. This extracted data is then combined with HD maps and dynamic vehicle information. The amalgamated data undergoes a process of local and global embedding, integrating information in both spatial and temporal domains. The final output of this process is the generation of a safe pathway for the autonomous vehicle, taking into account the comprehensive, fused data from the immediate and broader driving environment.

\subsection{Perception: Object Detection and Tracking via YOLOv8}

\begin{figure*}[t]
\centering
\includegraphics[width=0.8\textwidth]{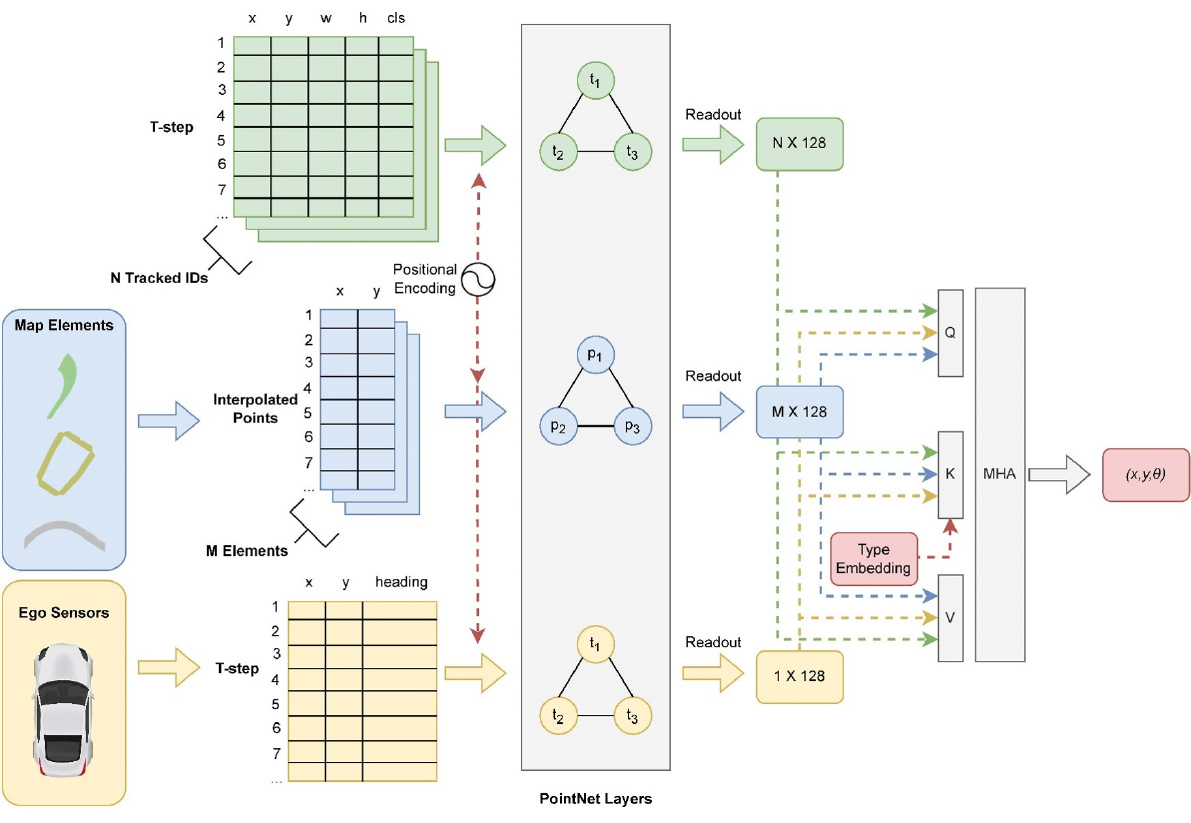}
\caption{The planning framework integrates data from perception systems, maps, and ego-motion sensors. This data, alongside positional encodings, is independently inputted into a fully-connected GNN. The output vectors from the GNN are then processed through a multi-head attention layer, which ultimately facilitates the generation of trajectory predictions.}
\label{fig_3}
\end{figure*}

In this phase, our objective is to construct a spatiotemporal feature matrix using multi-view camera inputs over past t time frames. We achieve this by organizing images from eight surrounding cameras into a 3x3 matrix, leaving the central position blank. To establish spatial correspondence, images from the left and right cameras are rotated 90 degrees counterclockwise and clockwise, respectively, and images from the three rear cameras are inverted. This arrangement is depicted in Figure \ref{fig_3}.

For the detection and tracking of objects, particularly those involved in traffic, we employ the YOLOv8 model pre-trained on the nuScenes dataset \cite{caesar_nuscenes_2020}. This model processes the composite images to generate bounding boxes and tracking numbers for each detected object. The tracking data is aligned with the past $t$ timesteps. The bounding boxes include class and location features, with location features comprising the pixel coordinates of the top-left corner, and the width and height of the bounding box. Thus, each object is represented by a five-dimensional feature vector. These bounding box features may also provide implicit information about the size of agents and distance to ego vehicle.

\subsection{HD Maps and Dynamic Features}

Beyond the visual data, our methodology integrates HD two-dimensional maps, which are extensively annotated with various semantic categories. These categories include roads, sidewalks, crosswalks, lanes, and traffic signals, and are sourced from the nuPlan dataset. The inclusion of these detailed maps is crucial for providing a comprehensive understanding of the vehicle's surrounding environment.

In addition, we incorporate dynamic data about the pose of the ego vehicle, collected by on-board sensors. This dynamic data encompasses critical information such as the ego vehicle's precise location, its yaw angle, and velocity. By combining these static map features with real-time dynamic information, we create a rich representation that significantly enhances the context available for our further path planning and decision-making algorithms.

\subsection{Path Planning}

Our path planning approach in urban environments emphasizes the importance of context-aware decision-making, taking into account the behaviors and interactions of various agents (vehicles, bicycles, pedestrians) and road infrastructure elements (crosswalks, lanes). To effectively manage this complexity, we adopt a graph-based modeling technique using GNNs. Inspired by the research of Scheel et al. and Gao et al., GNNs are chosen for their capability to represent data structures with nodes and edges, making them well-suited for capturing the intricate interactions and relationships within a graph-based framework.

We implement a dual-embedding strategy with GNNs to process and analyze the data. In the initial phase of this approach, we focus on embedding individual data points of each element. For each tracked object, the process involves embedding the bounding box information at each timestamp to create detailed feature vectors. These vectors encapsulate the spatiotemporal characteristics of the tracked objects, providing a comprehensive understanding of their movement and position over time. Similarly, the dynamic data from the ego vehicle, encompassing aspects such as location, velocity, and yaw angle, are embedded to capture temporal dynamics. This temporal embedding ensures that the changing states of the ego vehicle are accurately reflected in the model, allowing for more informed decision-making in real-time scenarios. Additionally, for the map elements, we employ a spatial embedding. This involves interpolating points within a Cartesian coordinate system, which allows us to create unique spatial feature representations for each map element. By doing so, we can effectively capture the static aspects of the environment, such as roads, lanes, and crosswalks, and integrate them into our path planning model.

The GNN architecture we deploy is a three-layer, fully connected network. This architecture does not rely on a pre-defined graph structure, allowing the network to autonomously learn and adapt the interrelationships among elements by modifying the weights of the edges. To generate the full feature representation of each element, we perform a summation readout of the node embeddings. Additionally, we introduce sinusoidal embedding to points of each object, enhancing the model's understanding of spatial and temporal order.

In the final phase of global embedding, our model employs a multi-head attention mechanism. This mechanism uses the ego feature as query and all features as key and value. This structure allows the model to synthesize a comprehensive overview from the perspective of the ego vehicle. We also integrate a type embedding into the key, enabling the model to differentiate and process various types of input effectively. The output vector from the attention layer is a rich interrelation feature, encapsulating a wealth of information about the ego vehicle and its surrounding environment. This vector essentially serves as an epitome of the situational context, integrating multiple data points into a cohesive whole. Processing the output of attention layer through a Multi-Layer Perceptron (MLP), the culmination of this process is the generation of a trajectory matrix (T × 3) for future T timestamps, alongside a sequence of waypoints, each characterized by an xy position and a yaw angle

In summary, our model is characterized by a streamlined structure that incorporates a state-of-the-art network architecture. This design choice was made with a focus on enabling fast inference and responsive decision making, which are critical requirements for real-world driving scenarios, and ensures that the model can efficiently process complex data inputs and make fast, accurate decisions, which are essential for navigating dynamic and unpredictable urban environments. This balance of simplicity and technological sophistication positions our model as a robust and effective solution for autonomous vehicle path planning.

\subsection{Loss Function}

In our network architecture, we have distinctly segmented the model into two parts: perception and planning. To optimize our resources and training efficiency, we approach the training of these two segments separately, each with its own tailored loss function.

For the perception module, we align our approach with the loss function utilized in YOLOv8. This decision is driven by the need for high accuracy in object detection and tracking, which is a critical component of the perception phase. The YOLOv8 loss function is designed to minimize errors in predicting the position, size, and class of the detected objects.

Turning to the planning module, our loss function strategy is inspired by the framework proposed in the research of Scheel et al. This approach is multifaceted, incorporating not only an L1 loss for imitation learning but also auxiliary loss terms that emphasize comfort and safety. The L1 loss component is crucial for enabling the model to mimic expert driving behavior, thereby learning path planning strategies. The auxiliary loss terms add an additional layer of sophistication to our model, as they allow us to fine-tune the model’s performance with respect to passenger comfort and vehicular safety. These terms ensure that the paths generated are not only optimal in terms of reaching the destination but are also smooth and safe, accounting for various dynamic factors such as traffic conditions, pedestrian movements, and road quality.

\section{Experiments}

This section delineates the dataset selection and the training methodology employed in developing our machine learning-based path planning model.

\begin{figure}[h]
\centering
\includegraphics[width=0.5
\textwidth]{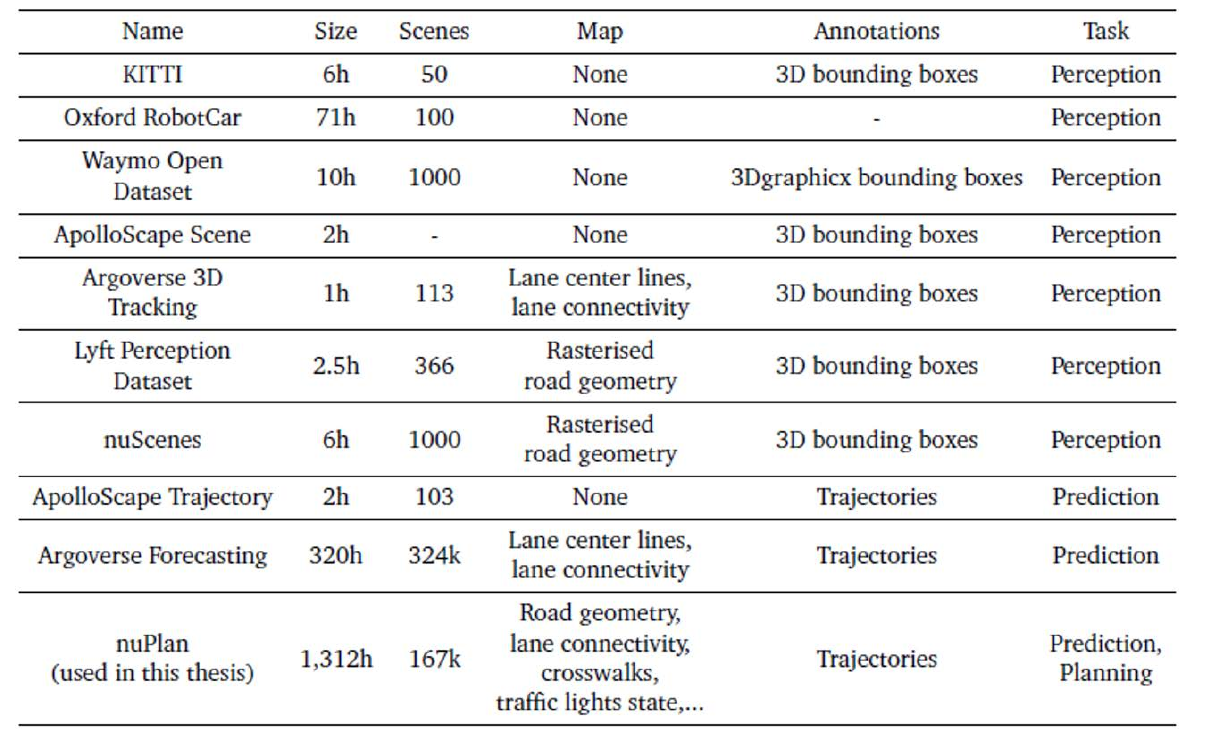}
\caption{Available autonomous driving datasets.}
\label{fig_4}
\end{figure}

\subsection{Dataset}

After carefully considering all available data sets in Figure \ref{fig_4}, we select different data sets for perception and planning tasks. For the training of our object detection system, we utilized the nuImages dataset, a comprehensive collection specifically tailored for AD applications. The nuImages dataset \cite{caesar_nuscenes_2020} comprises approximately 93,000 accurately annotated 2D images derived from challenging real-world driving scenes, encompassing 23 distinct classes. Our model is trained to detect 10 of the most prevalent traffic participant and infrastructure categories, including barriers, bicycles, buses, cars, construction vehicles, motorcycles, pedestrians, traffic cones, trailers, and trucks. We trained the YOLOv8 detector for 50 epochs on this dataset, starting from a pretrained YOLOv8x model, to achieve exceptional object detection performance.

Additionally, for training our path planning model, we utilized the nuPlan dataset \cite{houston_one_2021}. As the first large-scale planning benchmark in the AD field, nuPlan offers over 120 hours of human driving sensor data from diverse urban centers like Boston, Pittsburgh, Las Vegas, and Singapore, known for their complex traffic conditions. The diversity in urban landscapes, vegetation, architectural styles, and driving patterns across these cities enhances the model's generalization ability. The dataset includes vision data from eight calibrated cameras, detailed 2D high-definition maps with various semantic annotations (roads, sidewalks, crosswalks, lanes, traffic lights, etc.), and dynamic data from ego sensors, such as GPS and IMU, providing location, velocity, and acceleration measurements. A key feature of nuPlan is its scenario mining framework, which offers a range of challenging corner-case scenarios crucial for addressing the long-tail problem in AD algorithms. nuPlan stands out in its provision of extensive driving logs, raw data, and a well-suited simulator, making it particularly valuable for our path planning training, simulation, and development.

Owing to resource constraints, we trained our planner on a subset of the nuPlan dataset, comprising 72 logs and over 60 hours of driving data. The dataset was divided into training and validation sets in an 8:2 ratio. The trained planner was then tested in the nuPlan simulator using scenarios excluded from the mini dataset. Our model's performance was evaluated across five key criteria:

\begin{itemize}
    \item[$\bullet$] Traffic Rule Violation: This includes metrics such as collision rates, off-road trajectory incidences, time gaps to lead agents, time-to-collision, and relative velocities during passing maneuvers.
    \item[$\bullet$] Human Driving Similarity: We assessed maneuver satisfaction in comparison to human driving, focusing on longitudinal velocity error, stop position error, and lateral position error. We also compared jerk/acceleration levels to those typical of human drivers.
    \item[$\bullet$] Vehicle Dynamics: This aspect quantified rider comfort and trajectory feasibility, measuring jerk, acceleration, steering rate, and vehicle oscillation. We also evaluated the compliance of these factors with predefined limits.
    \item[$\bullet$] Goal Achievement: We measured the efficiency of route progress towards designated waypoints using L2 distance.
    \item[$\bullet$] Scenario-Specific Metrics: Tailored metrics were applied to assess performance in specific scenarios presented in the test set.
\end{itemize}

\subsection{Implementation Details}

In our experimental setup, we began by processing input images from eight channels. These images were stitched together and resized to a resolution of 640x640 pixels. A sequence of these stitched images, spanning across multiple timestamps 't', was then fed into a pretrained YOLO model. This process led to the construction of a 3D detection result matrix. The first dimension of this matrix corresponds to the identification of detected objects, tracked using their unique IDs. The second dimension represents the time dimension, capturing the states of tracked objects at each timestamp. The third dimension, or the feature dimension, includes data on bounding boxes and class information of the detected objects. The HD maps features were similarly constructed into a 3D matrix. The first dimension of this matrix delineates different road elements, while the second dimension represents interpolated points that outline the shapes of these elements. The feature dimension, constituting the third dimension, includes the Euclidean coordinates of these points. Additionally, the dynamic information of the ego vehicle was compiled into a 2D matrix, with the first dimension capturing timestamps and the second dimension encompassing features such as location, yaw angle, and velocity.

Subsequently, we expanded the first dimension of the ego feature matrix and stack it with other feature matrices. This led to the formation of a unified matrix with three dimensions: elements, points, and features. This comprehensive matrix was then inputted into a Multilayer Perceptron (MLP) to transform the feature vectors into a length of 256, aiming to extract rich information from the data. Following this, the data underwent processing in a 3-layer GNN, resulting in a descriptor of size 256 for each element. The process continued with the application of a single layer of scaled dot-product attention, integrated with type embedding, to obtain a feature vector of size 128. The final step involved another MLP layer, which projected the output into a specific shape representing planned trajectories for future $T-t$ timestamps. This output was utilized to compute the L1 loss against the ground truth of future trajectories. Utilizing both L1 loss and auxiliary loss, we conducted gradient descent to optimize the parameters of our path planning mode

\section{Conclusion}

In this paper, we proposed a vision-centric, end-to-end AD framework, utilizing the YOLO algorithm for object perception and an attention-based GNN for path planning. This approach marks a stride in AD, especially for navigating complex urban environments. Our method integrates high-resolution camera imagery and HD maps with dynamic vehicle data. This integration is crucial for accurately processing spatial and temporal information, essential in unpredictable urban settings. Our dual embedding system, combining local and global data, effectively enhances path planning accuracy. The model's performance was evaluated using the nuPlan simulator across various urban driving scenarios. The results demonstrated our model's effectiveness compared to existing vision-centric AD strategies and baselines, indicating its potential to significantly improve AD technologies.

Key to our approach is the model's capability to emulate human-driver decision-making in complex scenarios, as proven in the nuPlan challenge. This ability, combined with the integration of advanced machine learning techniques in perception and planning, suggests a path towards more robust, adaptable, and safer autonomous vehicles, particularly for higher automation levels.

\bibliographystyle{plain}
\bibliography{paper}

\begin{thebibliography}{10}

\bibitem{caesar_nuscenes_2020}
Holger Caesar, Varun Bankiti, Alex~H. Lang, Sourabh Vora, Venice~Erin Liong, Qiang Xu, Anush Krishnan, Yu~Pan, Giancarlo Baldan, and Oscar Beijbom.
\newblock {nuScenes}: A multimodal dataset for autonomous driving.
\newblock pages 11621--11631.

\bibitem{codevilla_end--end_2018}
Felipe Codevilla, Matthias Muller, Antonio Lopez, Vladlen Koltun, and Alexey Dosovitskiy.
\newblock End-to-end driving via conditional imitation learning.
\newblock In {\em 2018 {IEEE} International Conference on Robotics and Automation ({ICRA})}, pages 4693--4700. {IEEE}.

\bibitem{gao_vectornet_2020}
Jiyang Gao, Chen Sun, Hang Zhao, Yi~Shen, Dragomir Anguelov, Congcong Li, and Cordelia Schmid.
\newblock {VectorNet}: Encoding {HD} maps and agent dynamics from vectorized representation.

\bibitem{girshick_rich_2014}
Ross Girshick, Jeff Donahue, Trevor Darrell, and Jitendra Malik.
\newblock Rich feature hierarchies for accurate object detection and semantic segmentation.
\newblock pages 580--587.

\bibitem{gong_real-time_2023}
Liang Gong, Yingxin Wu, Bishu Gao, Yefeng Sun, Xinyi Le, and Chengliang Liu.
\newblock Real-time dynamic planning and tracking control of auto-docking for efficient wireless charging.
\newblock 8(3):2123--2134.

\bibitem{grigorescu_survey_2020}
Sorin Grigorescu, Bogdan Trasnea, Tiberiu Cocias, and Gigel Macesanu.
\newblock A survey of deep learning techniques for autonomous driving.
\newblock 37(3):362--386.
\newblock \_eprint: https://onlinelibrary.wiley.com/doi/pdf/10.1002/rob.21918.

\bibitem{hasirlioglu_test_2016}
Sinan Hasirlioglu, Alexander Kamann, Igor Doric, and Thomas Brandmeier.
\newblock Test methodology for rain influence on automotive surround sensors.
\newblock In {\em 2016 {IEEE} 19th International Conference on Intelligent Transportation Systems ({ITSC})}, pages 2242--2247.
\newblock {ISSN}: 2153-0017.

\bibitem{he_mask_2017}
Kaiming He, Georgia Gkioxari, Piotr Dollar, and Ross Girshick.
\newblock Mask r-{CNN}.
\newblock pages 2961--2969.

\bibitem{houston_one_2021}
John Houston, Guido Zuidhof, Luca Bergamini, Yawei Ye, Long Chen, Ashesh Jain, Sammy Omari, Vladimir Iglovikov, and Peter Ondruska.
\newblock One thousand and one hours: Self-driving motion prediction dataset.
\newblock In {\em Proceedings of the 2020 Conference on Robot Learning}, pages 409--418. {PMLR}.
\newblock {ISSN}: 2640-3498.

\bibitem{hu_st-p3_2022}
Shengchao Hu, Li~Chen, Penghao Wu, Hongyang Li, Junchi Yan, and Dacheng Tao.
\newblock {ST}-p3: End-to-end vision-based autonomous driving via spatial-temporal feature learning.

\bibitem{hu_planning-oriented_2023}
Yihan Hu, Jiazhi Yang, Li~Chen, Keyu Li, Chonghao Sima, Xizhou Zhu, Siqi Chai, Senyao Du, Tianwei Lin, Wenhai Wang, Lewei Lu, Xiaosong Jia, Qiang Liu, Jifeng Dai, Yu~Qiao, and Hongyang Li.
\newblock Planning-oriented autonomous driving.

\bibitem{jain_autonomy_2021}
Ashesh Jain, Luca Del~Pero, Hugo Grimmett, and Peter Ondruska.
\newblock Autonomy 2.0: Why is self-driving always 5 years away?

\bibitem{levine_continuous_2012}
Sergey Levine and Vladlen Koltun.
\newblock Continuous inverse optimal control with locally optimal examples.

\bibitem{lin_focal_2017}
Tsung-Yi Lin, Priya Goyal, Ross Girshick, Kaiming He, and Piotr Dollar.
\newblock Focal loss for dense object detection.
\newblock pages 2980--2988.

\bibitem{liu_yolo-bev_2023}
Chang Liu, Liguo Zhou, Yanliang Huang, and Alois Knoll.
\newblock {YOLO}-{BEV}: Generating bird's-eye view in the same way as 2d object detection.

\bibitem{liu_ssd_2016}
Wei Liu, Dragomir Anguelov, Dumitru Erhan, Christian Szegedy, Scott Reed, Cheng-Yang Fu, and Alexander~C. Berg.
\newblock {SSD}: Single shot {MultiBox} detector.
\newblock In Bastian Leibe, Jiri Matas, Nicu Sebe, and Max Welling, editors, {\em Computer Vision – {ECCV} 2016}, Lecture Notes in Computer Science, pages 21--37. Springer International Publishing.

\bibitem{redmon_you_2016}
Joseph Redmon, Santosh Divvala, Ross Girshick, and Ali Farhadi.
\newblock You only look once: Unified, real-time object detection.
\newblock pages 779--788.

\bibitem{scheel_urban_2021}
Oliver Scheel, Luca Bergamini, Maciej Wołczyk, Błażej Osiński, and Peter Ondruska.
\newblock Urban driver: Learning to drive from real-world demonstrations using policy gradients.

\bibitem{schwarz_laser-induced_2017}
Bastian Schwarz, Gunnar Ritt, Michael Koerber, and Bernd Eberle.
\newblock Laser-induced damage threshold of camera sensors and micro-optoelectromechanical systems.
\newblock 56(3):034108.
\newblock Publisher: {SPIE}.

\bibitem{teng_motion_2023}
Siyu Teng, Xuemin Hu, Peng Deng, Bai Li, Yuchen Li, Yunfeng Ai, Dongsheng Yang, Lingxi Li, Zhe Xuanyuan, Fenghua Zhu, and Long Chen.
\newblock Motion planning for autonomous driving: The state of the art and future perspectives.
\newblock 8(6):3692--3711.
\newblock Conference Name: {IEEE} Transactions on Intelligent Vehicles.

\bibitem{vitelli_safetynet_2021}
Matt Vitelli, Yan Chang, Yawei Ye, Maciej Wołczyk, Błażej Osiński, Moritz Niendorf, Hugo Grimmett, Qiangui Huang, Ashesh Jain, and Peter Ondruska.
\newblock {SafetyNet}: Safe planning for real-world self-driving vehicles using machine-learned policies.

\bibitem{zhang_attention-based_2023}
Zhengming Zhang, Renran Tian, Rini Sherony, Joshua Domeyer, and Zhengming Ding.
\newblock Attention-based interrelation modeling for explainable automated driving.
\newblock 8(2):1564--1573.

\bibitem{ziegler_navigating_2008}
J.~Ziegler, M.~Werling, and J.~Schroder.
\newblock Navigating car-like robots in unstructured environments using an obstacle sensitive cost function.
\newblock In {\em 2008 {IEEE} Intelligent Vehicles Symposium}, pages 787--791.
\newblock {ISSN}: 1931-0587.

\end{thebibliography}

\end{document}